# Contemporary Amharic Corpus: Automatically Morpho-Syntactically Tagged Amharic Corpus

**Andargachew Mekonnen Gezmu**
Otto-von-Guericke-Universität
andargachew.gezmu@ovgu.de

**Binyam Ephrem Seyoum**
Addis Ababa University
binyam.ephrem@aau.edu.et

**Michael Gasser**
Indiana University
gasser@cs.indiana.edu

**Andreas Nürnberger**
Otto-von-Guericke-Universität
andreas.nuernberger@ovgu.de

## Abstract

We introduced the contemporary Amharic corpus, which is automatically tagged for morpho-syntactic information. Texts are collected from 25,199 documents from different domains and about 24 million orthographic words are tokenized. Since it is partly a web corpus, we made some automatic spelling error correction. We have also modified the existing morphological analyzer, HornMorpho, to use it for the automatic tagging.

## 1  Introduction

Amharic is a Semitic language that serves as working language of the Federal Government of Ethiopia. Next to Arabic, it is the most spoken Semitic language. Though it plays several roles, it is considered one of the less-resourced languages. This is because it lacks basic tools and resources for carrying out natural language processing research and application.

One of the main hurdles in natural language processing of Amharic is the absence of sizable, clean and properly tagged corpora. Since Amharic is morphologically rich and the boundary of the syntactic word in an orthographic word is unclear in most cases. Even if it is possible to use some techniques of collecting a big corpus from the web, the basic natural language tasks like POS tagging or parsing would be challenging if we base our analysis on the orthographic words. In Amharic orthographic words represent different linguistic units. It is possible for an orthographic word to represent a phrase, or a clause or a sentence. This is because some syntactic words like prepositions, conjunctions, axillaries, etc can be attached to other lexical categories. For proper tagging and syntactic analysis then these morphemes should be separated from their phonological host. In this project, we made an effort to collect, segment, and tag text documents from various sources.

## 2  Related Work

Even though Amharic is a less-resourced language, there are some corpus collections by different initiatives. Since the introduction of the notion of considering the web as a corpus, which is motivated for practical reasons of getting larger data with open access and low cost, there have been corpora of different sizes for Amharic. In this section, we describe three corpora: Walta Information Center (WIC), HaBit (amWaC17), and An Crúbadán.

The WIC corpus is a medium-sized corpus of 210,000 tokens collected from 1065 Amharic news documents (Demeke and Getachew, 2006). The corpus is manually annotated for POS tags. The number of tag-sets, which is proposed based on the orthographic word, is 31. They use compound tag-sets for those words with prepositions and conjunctions. Even though they distinguish some subclasses of a given category (like noun and verb), the distinction no longer exists when these forms attach prepositions and/or conjunctions. However, they propose independent tags for both prepositions and conjunctions without favoring segmentation of these clitics. In the annotation process, annotators were given training and guidelines. Each annotator was given a different news article. They were asked to write the tags by hand and the annotation was given to a typist to insert the handwritten tag into the document. In such a process, there may be some loss of information which could be attributed to the inconsistencies of the annotation (Gambäck et al., 2009; Tachbelie and Menzel, 2009; Yimam et al., 2014). In the WIC report, they did not provide the annotation consistencies measurement and method of keeping consistency during the process of the annotation. According to the recent report of Rychlý and Suchomel (2016), the efficiency of an automatic tagger developed or trained on this corpus is 87.4%. Since this corpus can be accessible to most NLP researchers on Amharic, it is used to train a stemmer (Argaw and Asker, 2007), Named Entity recognition (Alemu, 2013) and a chunker (Ibrahim and Assabie, 2014).

The HaBit corpus, called amWaC17 (Amharic 'Web as Corpus', the year 2017) is another web corpus, which was developed by crawling using SpiderLing (Rychlý and Suchomel, 2016). Most of the crawling was done in August 2013, October 2015, and January 2016. The current version includes a web corpus crawled from May to September 2017 and consists of 30.5 million tokens collected from 75,509 documents. The corpus is cleaned and tagged for POS using a TreeTagger trained on WIC. They have used the same tag-sets used in WIC and followed the same principles as WIC in tagging tokens.

The An Crúbadán corpus was developed under the project called corpus building for under-resourced languages. The initiative aimed at the creation of text corpora for a large number of under-resourced languages by crawling the web (Scannell, 2007). The project collected written corpora for more than 2000 languages. Amharic was one of the languages to be included in this project. The Amharic corpus consists of 16,970,855 words crawled from 9999 documents. The major sources used are Amharic Wikipedia, Universal Declaration of Human Right, and JW.org. This corpus is a list of words with their frequencies.

In the three corpora mentioned above, the written or the orthographic form is considered as a word. Syntactic words like prepositions, conjunctions, and articles are not separately treated. In other words, clitics were not segmented from their host. In addition, though it is possible to collect authentic data entry from the web, such sources are inaccurate. As Amharic is not standardized, in these sources one may face lots of variation and expect to find misspellings, typographical errors, and grammatical errors. This calls for manual or automatic editing.

## 3 Data Collection and Preprocessing

### 3.1 Data Collection

The Contemporary Amharic Corpus (CACO) is collected from archives of various sources which are edited and made available for the public. All of the documents are written in modern or contemporary Amharic. Table 1 summarizes documents used in the corpus.

| Type of Documents | Titles |
|---|---|
| Newspapers | አዲስ አድማስ, አዲስ ዘመን, ሪፖርተር, ነጋሪት ጋዜጣ |
| News articles | Ethiopian News Agency, Global Voices |
| Magazines | ንቁ, መጠበቂያ ግንብ |
| Fictions | የልምዝት, ግርሽ, ልጅነት ተመልሶ አይመጣም, የአመጽ ኑዛዜ, የቅናት ዛር, አግዐዚ |
| Historic novel | አሉላ አባነጋ, ማዕበል የአብዮቱ ማግስት, የማይጨው ቁስለኛ, የታንጉት ሚስጢር |
| Short stories | የዓለም መስታወት, የቡና ቤት ስዕሎችና ሌሎችም ወጎች |
| History books | አጭር የኢትዮጲያ ታሪክ, ዳግማዊ አጼ ምኒልክ, ዳግማዊ ምኒልክ, የአቴጌ ጣይቱ ብጡል (፲፰፻፶፬ - ፲፱፻፩) አጭር የሕይወት ታሪክ, ከወለወል እስከ ማይጨው |
| Politics book | ማርክሲዝምና የቋንቋ ችግሮች, ሞት የማን ነው |
| Children's book | ፒኖኪዮ, ውድድር |
| Amharic Bible[1] | አዲስ ዓለም ትርጉም መጽሐፍ ቅዱስ |

Table 1: Data sources for CACO.

---
[1] We used the New World Translation of the Bible which is translated into the contemporary (not archaic) Amharic.

In total 25,199 documents from these sources are preprocessed and tagged. The preprocessing and tagging tasks are discussed in the subsequent sections.

### 3.2 Preprocessing

The preprocessing of the documents involves spelling correction, normalization of punctuation marks, and sentence extraction from paragraphs. In the documents, different types of misspellings are observed. Misspellings result from missed out spaces (e.g., አንዳንድየህክምናተቋማትናባለሙያዎቻቸው), replacing letters with visually similar characters (e.g., ቆጠሬ for ቆጠረ), and typographical errors. In addition, four Amharic phonemes have one or more homophonic character representations. The homophonic characters are commonly observed to be used interchangeably, which can be considered real-word misspellings. To correct the misspellings, we used the spelling corrector developed by Gezmu et al. (2018). Mainly the spelling corrector is employed to correct the first two types of spelling errors. As intensive manual intervention is needed to select the correct spelling from the plausible suggestions for typographical errors, in the current version of the corpus we have not corrected the typographical errors. To deal with the problems of real-word spelling errors resulting from homophonic letters, we adhere to the Ethiopian Languages Academy (ELA) spelling reform (ELA, 1970; Aklilu, 2004). Following their reform, homophonic characters are replaced with their common forms; ሐ and ኀ are replaced with ሀ, ሠ with ሰ, ዐ with አ, and ፀ with ጸ.

The other issue that needs attention is normalization of punctuation marks. Different styles of punctuation marks have been used in the documents. For instance, for double quotation mark two single quotation marks, ", ", ‹‹, ››, ``, ", « or » are used. Thus, normalization of punctuation marks is a non-trivial matter. We normalized all types of double quotes by ", all single quotes by ', question marks (e.g., ? and ፧) by ?, word separators (e.g., ፡ and ፥) by plain space, full stops (e.g., ። and ።) by ።, exclamation marks (e.g., ! and ፨) by !, hyphens (e.g., :-, and ፦) by ፦, and commas (e.g., ፣ and ÷) by ፣.

From the collected documents, sentences are identified and extracted by their boundaries—either double colon-like symbols (።) or question marks (?)—and are tokenized based on the orthographic-word boundary, a white space.

About 1.6 million sentences are collected from the documents. When the sentences are tokenized by plain space as a word boundary, around 24 million orthographic words are found. We build a 3-gram language model using the corpus. The corpus statistics are given in Table 2.

| Elements | Numbers |
|---|---|
| Sentences | 1,605,452 |
| Tokens | 24,049,484 |
| Unigrams | 919,407 |
| Bigrams | 9,170,309 |
| Trigrams | 16,033,318 |

Table 2: Statistical information for CACO corpus.

## 4 Segmentation and Tagging

### 4.1 Amharic Orthography

The Amharic script is known as Ge'ez or Ethiopic. The system allows representing a consonant and a vowel together as a symbol. Each symbol represents a CV syllable. However, a syllable in Amharic may have a CVC structure. It is worth taking into account that the writing system does not mark gemination. Gemination is phonemic; it can bring meaning distinctions and in some cases can also convey grammatical information.

As in other languages, Amharic texts shows variation in writing compound words. They can be written in three ways; with a space between the compound words (e.g., ሆደ ሰፊ /hodə səffi/ "patient", ምግብ ቤት /mɨgɨb bet/ "restaurant"), separated by a hyphen (e.g., ስነ-ልሳን /sɨnə-lɨsan/ "linguistics", ስነ-ምግባር /sɨnə-mɨgɨbar/ "behavior") or written as a single word (e.g., ቤተሰብ /betəsəb/ "family", ምድረበዳ /mɨdrəbəda/ "desert").

Orthographic words, which are separated by whitespace (semicolon in old documents), may be coupled with functional words like a preposition, a conjunction or auxiliaries. In most cases, an orthographic word may function as a phrase (e.g., ከቤትዋ /kəbətwa/ "from her house"), a clause (e.g., የመጣው /jəmmət'ʕaw/ "the one who came"), or even a sentence (e.g., አልበላችም /ʔalbəllatʃtʃɨm/ "She did not eat."). In addition, the writing system allows people to write in a phonemic form (the abstract form or what one intends to say) or in a phonetic form (what is actually uttered). As a result, we have different forms for a given word.

All the above features need to be addressed in processing Amharic texts. Some of the above problems call for standardization efforts to be made whereas others are due to the decision to write what is in mind and what is actually produced. Furthermore, for proper syntactic analyses clitics could be segmented. We used an existing morphological analyzer with some improvements to perform the possible segmentation and tagging.

### 4.2 Proposed Tag-sets

As we have indicated in Section 2, most works on POS tags in Amharic are based on orthographic words. In such approach, they proposed compound tagsets for words with other syntactic words. Tagged corpora following this approach may not be used for syntactic analyses. For syntactic analysis, we need information about the structure within a phrase, a clause and a sentence. Tagging with a bundle of tagset will hide such syntactic information.

Since one of the main objectives of tagging task is to provide lexical information that can be employed for syntactic analysis like parsing, we suggest that syntactic words should be the unit of analysis for tagging rather than orthographic words. In light of this, we can follow what is proposed by Seyoum et al. (2018). In this work syntactic word is used for the analysis in favor of a lexicalist syntactic view. They suggested that clitics should be segmented as a pre-processing step. Since manual segmentation is very costly and time taking we opted for doing morphological segmentation using an existing tool that could be closer to our intention. For this purpose, we have found HornMorpho to be a possible candidate with some improvement. Although this tagset has not yet been fully integrated into HornMorpho, we use it for analysis since no other analyzer has comparable coverage.

### 4.3 HornMorpho Analysis and Limitations

HornMorpho (Gasser, 2011) is a rule-based system for morphological analysis and generation. As in many other modern computational morphological systems, it takes the form of a cascade of composed finite-state transducers (FSTs) that implement a lexicon of roots and morphemes, morphotactics, and alternation rules governing phonological or orthographic changes at morpheme boundaries (Beesley and Karttunen, 2003). To handle the complex long-distance dependencies between morphemes that characterize the morphology of Amharic, the transducers in HornMorpho are weighted with feature structures, an approach originally introduced by Amtrup (2003). During transduction, the feature structure weights on arcs that are traversed are unified, resulting in transducer output that includes an accumulated feature structure as well as a character string. In this way, the transducers retain a memory of where they have been, allowing the system to implement agreement and co-occurrence constraints and to output characters based on morphemes encountered earlier in the word.

As an example, consider the imperfective forms of Amharic verbs, which, as in other Semitic languages, include subject person/gender/number agreement prefixes as well as suffixes. The prefix t- encodes either third person singular feminine or second person singular masculine, singular feminine or plural. The suffix following an imperfective stem, or absence thereof, may disambiguate the prefix. Once it has seen the prefix, the HornMorpho transducer "remembers" the set of possible subject person, gender, and number features that are consistent with the input so far. When it reaches the suffix (or finds none), it can reduce the set of possible features (or reject the word if an inconsistent suffix is found). So in the word ትፈልጊያለሽ /tɨfəllɨgijalləʃ/ "you (sing. fem.) want", t - fəllɨg - i - alləʃ, the presence of the suffix -i following the stem -fəllɨg- tells HornMorpho to constrain the subject person and number to second person singular feminine; without the proper prefix, HornMorpho would reject the word at this point.

The major focus in the development of HornMorpho was on lexical words not on function words. More specifically, it analyzes nouns and verbs. Since Amharic adjectives behave like nouns, HornMorpho does not distinguish between adjectives and nouns. In addition, compound words and light verb constructions (in version 2.5) are not handled. It also gives wrong results when it gets words with more than two clitics attached to a word. For instance, በበኩላቸው /bəbbəkkulaʧʧəw/ "from his/her side or perspective", (this word occurs 4,293 times in amWaC17). For this word, the system guessed fifteen analyses. Furthermore, since clitics are not considered as separate words, the system does not give any analysis for clitics. Therefore, in order to use this tool for tagging a big corpus, we need to improve the coverage of the tool.

### 4.4 Improvement to HornMorpho

For the purposes of this paper, we modified HornMorpho in several ways. Since the major focus of HornMorpho is to give analysis for nouns and verbs, it does not give morphological analysis for syntactic words. However, as a result of the improvement, it distinguishes more parts-of-speech (verbs, nouns, adjectives, adverbs, conjunctions, and adpositions). Morphological analysis for constructions like light verbs, compound nouns and verbs, are now included in the improved version. As a rule-based system, it has limited lexicon. The improved version, however, recognizes personal and place names. For our purpose, we have enhanced the coverage of the lexicon.

```
<s>
  <w pos="PRON" morphemes="be(prep)-{yh}" latin="bezihu">በዚሁ </w>
  <w pos="NM_PRS" morphemes="ye(gen)-{dereja}" lain="yedereja">
      የደረጃ </w>
  <w pos="N" morphemes="{zrzr}" latin="zrzr">ዝርዝር </w>
  <w pos="NADJ" morphemes="{'and}-m(cnj)" latin="'andm">አንድም </w>
  <w pos="N" morphemes="{'ityoP_yawi}" latin="'ityoPyawi">ኢትዮጵያዊ </w>
  <w pos="V" morphemes="al(neg1)-{ktt+te1a2_e3}(prf,recip,pas)-
      e(sb=3sm)-m(neg2)" latin="'altekatetem">አልተካተተም </w>
  <w pos="PUN">። </w>
</s>
```

Figure 1. An example sentence that is tagged with HornMorpho.

Figure 1 shows an example sentence from our corpus that is tagged by HornMorpho, በዚሁ የደረጃ ዝርዝር አንድም ኢትዮጵያዊ አልተካተተም ። /bəzihu jədərədʒa zirzir ʔandɪmm ʔitjopʼijawi ʔaltəkatətəm./ "No Ethiopian has been included in the ranking list.", transliterated as "bezihu yedereja zrzr 'andm 'ityoPyawi 'altekatetem ።". The HornMorpho segmenter puts the stem between {}, gives a POS label for the lexical word, and labels grammatical morphemes, which are mostly function words, that are separated by hyphens, with features/tags. For verbs, the stem is represented as a sequence of root consonants and a template consisting of root consonant positions, vowels and gemination characters. The root consonants and template are separated by a "+", the root consonant positions in the template are represented by numbers, and underscore indicates gemination. During analysis by HornMorpho, the Amharic text is transliterated to Latin using the System for Ethiopic Representation in ASCII (Firdyiwek and Yaqob, 1997).

### 5   Conclusion and Future Work

In this paper, we have introduced a new resource, CACO[2], a partly web corpus of Amharic. The corpus has been collected from different sources with different domains including newspapers, fiction, historical books, political books, short stories, children books and the Bible. These sources are believed to be well edited but still need automatic editing. The corpus consists of 24 million tokens from 25,199 documents. It is automatically tagged for POS using HornMorpho. We made some modifications to HornMorpho for the tagging. In doing so, we noticed that Amharic orthographic words should be properly segmented in a preprocessing stage. Currently, we have more words with unclassified tags because of the limitation of HornMorpho. Therefore, as future work, we plan to enhance HornMorpho for automatic tagging. In addition, we plan to increase the size of the data and apply our method to other existing corpora.

---

[2] It is freely available for research purposes at: http://dx.doi.org/10.24352/ub.ovgu-2018-144

## Acknowledgments

We would like to extend our sincere gratitude to the anonymous reviewers for their valuable feedback. Our thanks goes to Geez Frontier Foundation for some of the resources that we freely used. We would like also thank the University of Oslo for the partial support under a project called Linguistic Capacity Building-Tools for inclusive development of Ethiopia.